\documentclass[10pt,twocolumn,letterpaper]{article}

\usepackage[pagenumbers]{cvpr} 

\makeatletter
\@namedef{ver@everyshi.sty}{}
\makeatother
\usepackage{tikz}

\usepackage{graphicx}
\usepackage{amsmath}
\usepackage{amssymb}
\usepackage{booktabs}
\usepackage{array}
\makeatletter
\newcommand{\thickhline}{%
    \noalign {\ifnum 0=`}\fi \hrule height 2pt
    \futurelet \reserved@a \@xhline
}
\makeatother
\usepackage{pgfplots}

\usepackage[pagebackref,breaklinks,colorlinks]{hyperref}

\usepackage[capitalize]{cleveref}
\crefname{section}{Sec.}{Secs.}
\Crefname{section}{Section}{Sections}
\Crefname{table}{Table}{Tables}
\crefname{table}{Tab.}{Tabs.}

\def\cvprPaperID{2644} 
\def\confName{CVPR}
\def\confYear{2022}

\begin{document}

\title{RM-Depth: Unsupervised Learning of Recurrent Monocular Depth \\
in Dynamic Scenes\thanks{This research work is not for commercial use unless a prior arrangement
has been made with the author.}}

\author{Tak-Wai Hui\\
H-1 Research\\
{\tt\small eetwhui@gmail.com}
}
\maketitle

\begin{abstract}
Unsupervised methods have showed promising results on monocular depth estimation. However, the training data must be captured in scenes without moving objects. To push the envelope of accuracy, recent methods tend to increase their model parameters. In this paper, an unsupervised learning framework is proposed to jointly predict monocular depth and complete 3D motion including the motions of moving objects and camera. (1) Recurrent modulation units are used to adaptively and iteratively fuse encoder and decoder features. This not only improves the single-image depth inference but also does not overspend model parameters. (2) Instead of using a single set of filters for upsampling, multiple sets of filters are devised for the residual upsampling. This facilitates the learning of edge-preserving filters and leads to the improved performance. (3) A warping-based network is used to estimate a motion field of moving objects without using semantic priors. This breaks down the requirement of scene rigidity and allows to use general videos for the unsupervised learning. The motion field is further regularized by an outlier-aware training loss. Despite the depth model just uses a single image in test time and 2.97M parameters, it achieves state-of-the-art results on the KITTI and Cityscapes benchmarks.
\end{abstract}

\section{Introduction}
\label{sec:intro}

Visual perception is an important ability for human to understand and perceive the world. As a consequence, research work on scene geometry has attracted a lot of attention over several decades. This promotes the deployment of technology to numerous applications such as autonomous vehicle, interactive robot, virtual and augmented reality, and more. The problem of scene geometry generally involves estimating depth, camera motion\footnote{The words, ego-motion, camera motion and pose, are interchangeably used throughout the paper.}, and optical flow from an image sequence. The above computer vision tasks are often recovered together since they are coupled through geometric constraints~\cite{Ranjan19,Yin18}

Unlike depth from triangulation, single-image depth estimation is inherently ill-posed because there are multiple possible 3D points along each light ray towards the camera center. Convolutional neural networks have demonstrated the ability to exploit the relationship between a captured image and the corresponding scene depth~\cite{Eigen14,Laina16}. Recently, unsupervised methods~\cite{Godard17,Godard19,Ranjan19,Yin18,Zhou17} have achieved appealing performance than the early supervised counterparts. Their successes primarily rely on the use of the classical technique, structure from motion. Given at least two images, a novel view generated from an image will be consistent with another image in the pair if depth and camera motion are correctly estimated. However, this strictly requires scene rigidity, \textit{i.e.} the training data must be captured in scenes without moving objects other than the moving camera itself. To get rid of this requirement, stereo image sequences\cite{Godard17} and masking out dynamic objects~\cite{Ranjan19,Zhou17} are commonly adopted. Recent works tend to devise a multi-image approach~\cite{Watson21}, a large amount of model parameters~\cite{Guizilini20}, and semantic priors~\cite{Wang18} for improving the depth accuracy.

In this paper, an unsupervised learning framework of recurrent monocular depth, dubbed RM-Depth, is proposed to jointly predict depth, camera motion, and motion field of moving objects without requiring static scenes in the training data. RM-Depth requires neither a large number of parameters nor prior semantic information. Particularly, image pairs are used in training while only a single image is used for depth inference at test time. The contributions of this work are summarized as follows:

\begin{enumerate}
\item Recurrent modulation unit (RMU) – Fusion of feature maps across encoder and decoder often appears in depth estimation~\cite{Godard19,Zhou17}. In the proposed method, the decoder consists of RMUs. The fusion is iteratively refined by adaptive modulating the encoder features using the hidden state of RMU. This in turn improves the performance of single-image depth inference.

\item Residual upsampling – Conventionally, feature maps are upsampled using a single set of filters~\cite{Shi16,Zeiler11}. In this work, multiple sets of filters are proposed such that each set of them is specifically trained for upsampling some of the spectral components. This effectively improves upsampling along edges.

\item Motion field of moving objects – Besides camera motion, a 3D motion field of moving objects is recovered in a coarse-to-fine framework through a warping approach. This breaks down the scene rigidity assumption and allows to use general videos for the unsupervised learning. The unsupervised learning of motion field is further improved by introducing an outlier-aware regularization loss.
\end{enumerate}

With the above innovations, RM-Depth achieves state-of-the-art results on the KITTI and Cityscapes benchmarks. The depth model only requires 2.97M parameters, while it achieves 4.8 and 44 times reduction in model size comparing to the popular Monodepth2~\cite{Godard19} and PackNet~\cite{Guizilini20}, respectively. The project page of this paper is available at~\url{https://github.com/twhui/RM-Depth}.

\section{Related Work}
\subsection{Unsupervised Joint Learning of Depth and Egomotion}

\noindent\textbf{Depth from a Single Image.} A pioneer work from Zhou~\textit{et al.}~\cite{Zhou17} proposes an unsupervised learning framework for estimating depth and ego-motion. Based on~\cite{Zhou17}, Godard~\textit{et al.}~\cite{Godard19} introduce the per-pixel minimum reprojection loss, auto-masking of stationary pixels, and full-scale estimation loss for improving the unsupervised training. Mahjourian~\textit{et al.}~\cite{Mahjourian18} and Bian~\textit{et al.}~\cite{Bian19} explore the consistencies of 3D point clouds and depth maps across consecutive frames, respectively. Wang~\textit{et al.}~\cite{Wang18} devise to use direct visual odometry for pose estimation without requiring additional pose network. Recently, Guizilini~\textit{et al}.~\cite{Guizilini20} utilize 3D convolutions for packing and unpacking feature maps. Johnston ~\textit{et al.}~\cite{Johnston20} propose to estimate depth map using self-attention and disparity volume. Poggi~\textit{et al.}~\cite{Poggi20} impose depth uncertainty during unsupervised training. Unlike the prior works, RM-Depth introduces recurrent modulation units and residual upsampling in the depth model. The proposed components lead to the improved performance while the depth model just requires a very small number of parameters (2.97M).

The previous works recover rigid flow\footnote{A component of optical flow that is solely due to camera motion without considering moving objects in the scene.} through the projection of estimated scene depth, and hence moving objects in the scene cannot be taken into account. To recover full flow, Yin~\textit{et al.}~\cite{Yin18} propose to use a network cascade to estimate the residual flow accounting for moving objects. Ranjan~\textit{et al.}~\cite{Ranjan19} propose a framework that facilitates the coordinated trainings of depth, ego-motion, and optical flow. Their method reasons about segmenting a scene into static and moving regions. Chen~\textit{et al.}~\cite{Chen19} use a separated flow network and introduce an online optimization scheme. Different from the prior works, scene rigidity is not required in the training of RM-Depth. The motion network of RM-Depth recovers both camera and object motions. Therefore, full flow can be used for the unsupervised training. This in turn improves the performance on depth estimation.

\noindent\textbf{Depth from Multiple Images.} Wang~\textit{et al.}~\cite{Wang19} exploit the temporal correlation across consecutive frames by using convolutional long short-term memory (LSTM). Despite a 10-frame sequence is used, it just performs on par with Monodepth2 ~\cite{Godard19}. Li~\textit{et al.}~\cite{Li19} utilize the encoded feature resulting from a self-contained optical flow network as the input to each LSTM. However, their model requires 15 LSTM modules for a proper depth inference. Li~\textit{et al.}~\cite{Li20} propose a self-supervised online meta-learning that uses LSTM to aggregate spatial-temporal information in the past. Watson~\textit{et al.}~\cite{Watson21} propose a cost volume based approach to fuse temporal information. Unlike LSTM or GRU~\cite{Cho14}, the proposed recurrent modulation unit (RMU) uses features from a single static image as the input but not features from a time varying image sequence.

\begin{figure*}[ht]
\centering
\includegraphics[width=\textwidth]{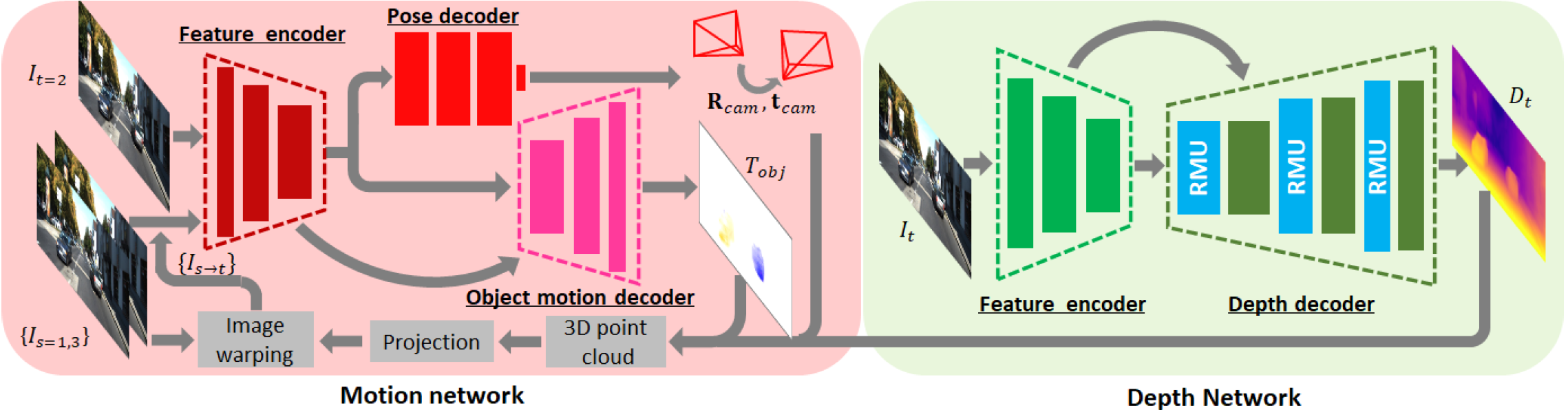}
\caption{An overview of the unsupervised learning framework. For brevity, a 3-level design is shown. Given an image sequence $\{I_{1}, I_{2}, I_{3}\}$, define $I_{t=2}$ as the target image and the rest $\{I_{s=1,3}\}$ as the source images. Depth map and motion field are estimated in a coarse-to-fine framework. For the motion network, $\{I_{s}\}$ are warped towards $I_{t}$ in accordance to the image projection computed by Eq.~\eqref{eq:novel view synthesis} using motion field $T_{obj}$ , camera pose $({\bf R}_{cam}, {\bf t}_{cam})$, and scene depth $D_{t}$. For the depth network, encoder and decoder features are adaptively and iteratively fused by RMUs. More details of the depth and motion networks are presented in Secs.~\ref{sec:recurrent depth network} and~\ref{sec:object motion}, respectively.}
\label{fig:overview}
\end{figure*}

\subsection{Unsupervised Joint Learning of Depth, Egomotion, and Object Motion}

Video data is often captured in scenes involving dynamic objects. Therefore, the assumption of scene rigidity is violated. Most of the prior works rely on additional segmentation labels to assist the unsupervised learning of object motion. With semantic prior, Casser~\textit{et al}.~\cite{Casser19} estimate the 3D motion of each dynamic object using a network similar to the one used for ego-motion. Gordon~\textit{et al.}~\cite{Gordon19} propose a network for estimating the motion field of moving objects. A pre-computed segmentation mask that pinpoints the locations of moving objects imposes regularization of the motion field. Li~\textit{et al.}~\cite{Li20} eliminate the use of semantic priors in~\cite{Gordon19} by introducing a sparsity loss. Gao~\textit{et al.} propose attentional CNN blocks to disentangle camera and object motion without semantic priors~\cite{Gao20}, but their experimental results are limited to the KITTI dataset. Lee~\textit{et al.}~\cite{Lee21} propose an instance-aware photometric and geometric consistency loss that imposes self-supervisory signals for static and moving object regions. RM-Depth estimates the motion field of moving objects without using semantic priors. A warping-based network is proposed for the motion field estimation. An outlier-aware training loss is further exploited for regularizing the motion field. Using the proposed innovations, RM-Depth outperforms the prior works.

\subsection{Unsupervised Learning of Depth Using Stereo Training Data}

The scene rigidity requirement limits unsupervised methods to use monocular data without involving dynamic objects in scenes. Since the left and right images in a stereo rig are captured simultaneously, stereo data provides an alternative option for the unsupervised training. Garg~\textit{et al.}~\cite{Garg16} propose to use the photometric difference between images in each stereo pair for governing the learning of monocular depth estimation. Godard~\textit{et al.}~\cite{Godard17} explore the consistency between the disparities produced relative to the left and right images. Zhan~\textit{et al.}~\cite{Zhan18} devise the temporal and spatial clues in stereo image sequences for improving the unsupervised training. Yang~\textit{et al.}~\cite{Yang20} aligns the illumination of the training images and model the photometric uncertainties of pixels on the input images.

\section{Depth from a Single Image}

Unsupervised learning of single-image depth estimation is often achieved by training two networks together~\cite{Godard19,Zhou17}. The primary (depth) network takes an image as the input and gradually predicts scene depth (up to a scale factor) with increasing spatial resolutions. The secondary (pose) network estimates camera motion for each image pair. Source frames in a given image sequence are warped towards the target frame by projecting the computed 3D point cloud to the target frame. The difference between the target and each synthesized frame is used as the driving force for the unsupervised training. In this work, an unsupervised learning framework RM-Depth is proposed for joint learning of depth, ego-motion, and object motion in general scenes. An overview of the learning framework is shown in Fig.~\ref{fig:overview}. In more details, the depth network utilizes Recurrent Modulation Units (RMU) to adaptively and iteratively combine encoder and decoder features (Sec.~\ref{sec:recurrent depth network}). Residual upsampling (Sec.~\ref{sec:residual upsampling}) is used to facilitate the learning of edge-aware filters. Furthermore, a 3D motion field of moving objects (Sec.~\ref{sec:object motion}) is recovered. As it will show later (Sec.~\ref{sec:experiments}), the proposed innovations lead to the improved depth accuracy despite not using any segmentation labels.

\subsection{Preliminaries}\label{sec:preliminaries}

\noindent\textbf{Perspective Projection.} Denote $O$ as the camera coordinate system associated with image $I$ and $\Omega \subset \mathbb{R}^{2}$ as the image domain. Suppose $D: \Omega \rightarrow \mathbb{R}$ is the depth map. A point ${\bf x} \in \Omega$ on $I$ is the image projection from a 3D point ${\bf p} \in \mathbb{R}^{3}$. Once $D({\bf x})$ (\textit{i.e}. z-coordinate of ${\bf p}$) is given, ${\bf p}$ can be recovered by back-projection of ${\bf x}$ as follows:
\begin{equation}
{\bf p} = D({\bf x}){\bf K}^{-1}\begin{pmatrix} {\bf x} & 1 \end{pmatrix}^\intercal,\label{eq:perspective projection}
\end{equation}
where ${\bf K}$ denotes a $3 \times3$ camera intrinsic matrix.

\vspace{+0.1em}\noindent\textbf{Novel View Synthesis.} Suppose an image sequence $\{I_{1},I_{2},...,I_{N}\}$ is given. In the following, subscripts $t$ and $s$ will be used to denote variables that are defined in the target and source views, respectively. Let’s consider one of the frames $I_{t}$ being the target view and the rest being the source views $I_{s}(1 \leq N, s \neq t)$. The transformation from $O_{t}$ to $O_{s}$ is governed by a $3\times3$ rotation matrix ${\bf R}$ and a 3D translation vector ${\bf t}$. Using Eq.~\eqref{eq:perspective projection}, the image projection of ${\bf p}_{t}$ onto $I_{s}$ is given by:
\begin{equation}
\begin{pmatrix} {\bf x}_{s} & 1 \end{pmatrix}^\intercal \cong {\bf K}\Bigl({\bf R}D_{t}({\bf x}_{t}){\bf K}^{-1} \begin{pmatrix} {\bf x}_{t} & 1 \end{pmatrix}^\intercal + {\bf t}\Bigr),\label{eq:novel view synthesis}
\end{equation}
where ``$\cong$'' denotes equality up to a positive scale factor and $D_{t}$ is the depth map at the target view. $I_{s}$ is warped towards $I_{t}$ to form a novel view $I_{s \rightarrow t}$ in accordance with the visual displacement (\textit{i.e.} optical flow) ${\bf x}_{s} - {\bf x}_{t}$.

\subsection{Recurrent Depth Network}\label{sec:recurrent depth network}

\begin{figure}[t]
\centering
\includegraphics[width=\linewidth]{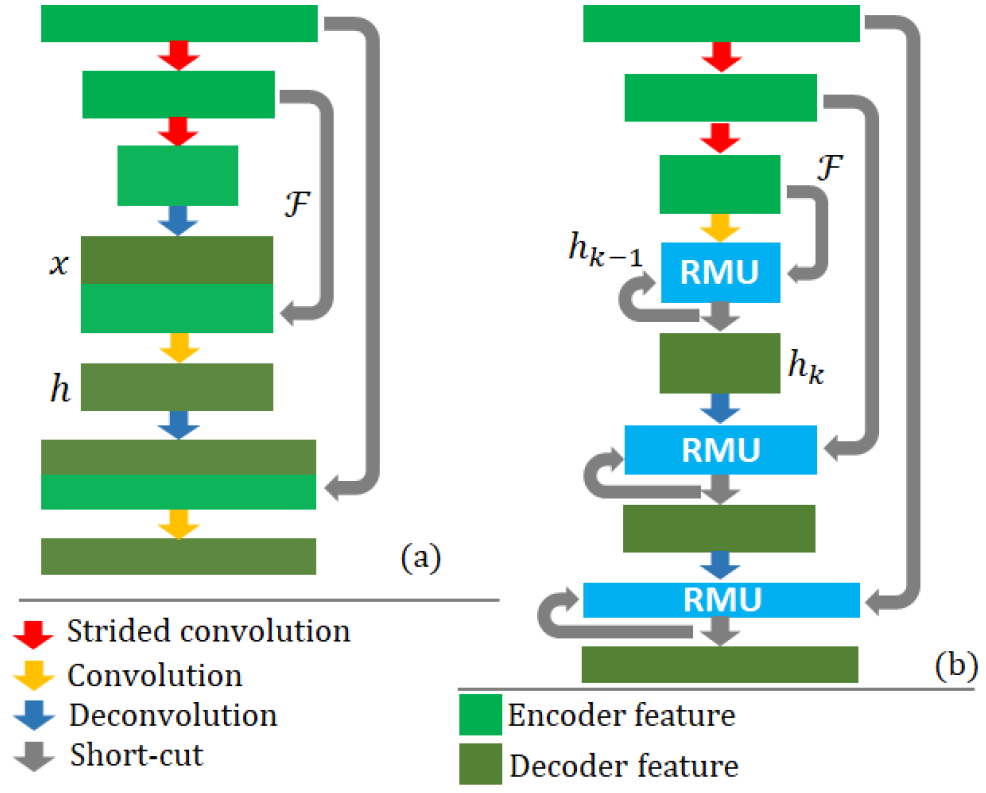}
\caption{The network architectures of different depth models: (a) Conventional method~\cite{Godard19,Zhou17} and (b) RMU-based model. For the ease of representation, a 3-level design is illustrated.}
\label{fig:different depth models}
\end{figure}

Top-down approach~\cite{Godard19,Zhou17} often adopts U-Net architecture~\cite{Ronneberger15} for depth inference. Fig.~\ref{fig:different depth models}\textcolor{red}{a} provides an overview of the network architecture. The upsampled decoder feature $x$ from the previous level is fused with the corresponding encoder feature $\mathcal{F}$ through a concatenation followed by a convolution layer. The feature fusion can be represented as follows:
\begin{equation}
h = \theta(\text{conv}([x,\mathcal{F}])),\label{eq:3}
\end{equation}
where ``$\theta$'' and ``conv'' represent an activation function and a convolution layer, respectively. Since the convolution kernels are fixed, the fusion cannot be adapted for different inputs. This limits the performance of depth inference.

It is desired to make the feature fusion to be adaptive. Intuitively, the decoder feature can be augmented with a modulated encoder feature. To this end, the encoder feature is adaptively transformed according to the current hidden state of the decoder. This is equivalent to change the feed-forward behavior of the encoder despite using the same input. Besides, recurrent CNN has been shown useful in improving network performance~\cite{Kim16}. Taking these inspirations, Recurrent Modulation Unit (RMU) is devised for dynamic and iterative feature fusion in the depth network. Fig.~\ref{fig:different depth models}\textcolor{red}{b} provides an overview of the proposed network. This design leads to the improved depth accuracy (Sec.\ref{sec:experiments}). In the following, when the operations are presented in a pyramid level, the same operations are applicable to other levels.

\begin{figure}[t]
\centering
\includegraphics[width=\linewidth]{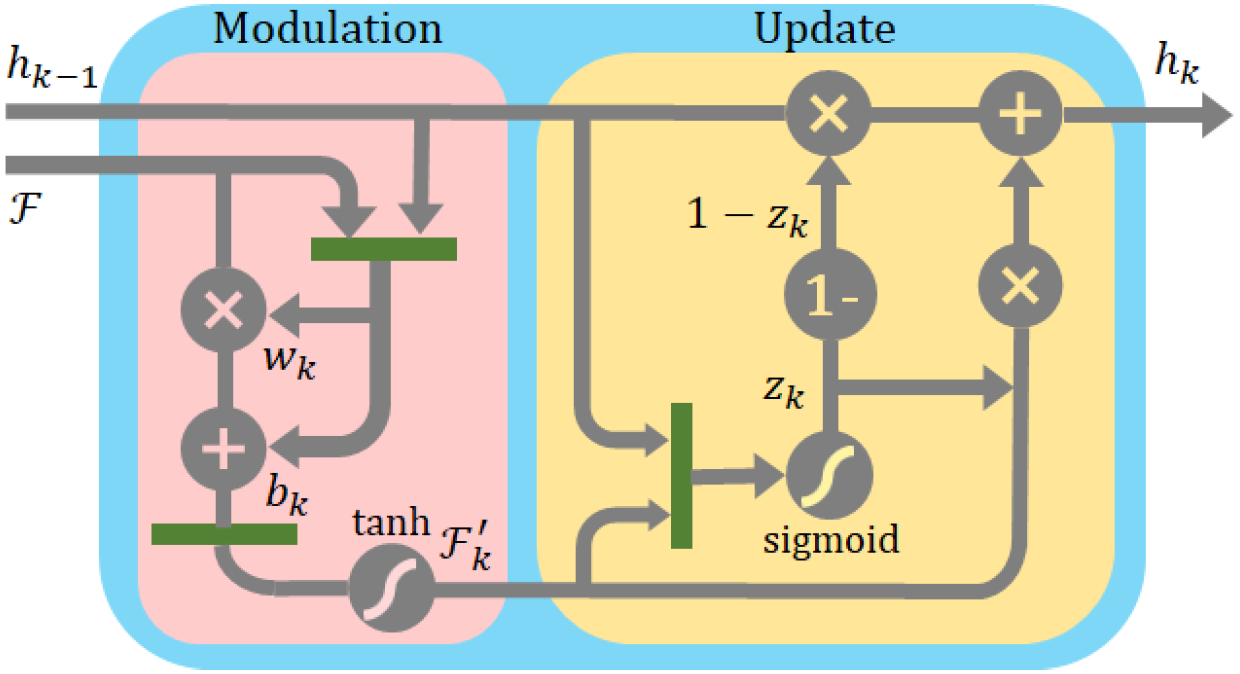}
\caption{The technical details of a RMU. At iteration $k$, the encoder feature $\mathcal{F}$ is modulated to $\mathcal{F}_{k}'$. The new hidden state $h_{k}$ is a weighted average between $\mathcal{F}_{k}'$ and the previous hidden state $h_{k-1}$ according to the element-wise adaptive scalar $z_{k}$.}
\label{fig:RMU}
\end{figure}

\vspace{+0.1em}\noindent\textbf{Recurrent Modulation Unit (RMU).} There are two components inside a RMU, namely \textit{modulation} and \textit{update}. Fig.~\ref{fig:RMU} shows the details. At iteration step $k$, the encoder feature $\mathcal{F}$ is adaptively modulated according to the previous fused feature $h_{k-1}$ (\textit{i.e.} the hidden state at iteration $k-1$) through an affine transformation\footnote{There could be other choices for the modulation function, affine transformation is selected because of its low computational complexity.} consisting of weight and bias terms $(w_{k}, b_{k})$ as follows (\textit{modulation phase}):
\begin{subequations}
\begin{align} 
w_{k},b_{k} &= \text{convs}([h_{k-1},\mathcal{F}]), \label{eq:4a} \\
\mathcal{F}_{k}' &= \text{tanh}(\text{conv}(w_{k} \odot \mathcal{F} + b_{k})), \label{eq:4b}
\end{align}
\end{subequations}
where ``convs'' and ``$\odot$'' denote convolutions and the Hadamard product, respectively. Eq~\eqref{eq:4a}. can be re-written to a residual form as $\text{conv}(\text{conv}(h_{k-1}) + conv(\mathcal{F}))$. Since $\mathcal{F}$ is fixed, the second term can be pre-computed to reduce the computational complexity. The hidden state $h_{k-1}$ is combined with the modulated encoder feature $F_{k}'$ for the feature fusion according to an element-wise adaptive scalar $z_{k}$ as follows (\textit{update phase}):
\begin{subequations}
\begin{align} 
z_{k} &= \sigma(\text{conv}([h_{k-1},\mathcal{F}_{k}'])), \label{eq:5a} \\
h_{k} &= (1-z_{k}) \odot h_{k-1} + z_{k} \odot \mathcal{F}_{k}', \label{eq:5b}
\end{align}
\end{subequations}
where ``$\sigma$'' denotes a sigmoid function. Particularly, the conventional feature fusion in Eq.~\eqref{eq:3} is static while the proposed feature fusion is both dynamic and iterative. 

Comparing to GRU~\cite{Cho14}, RMU uses features from a single static image as the input but not features from a time-varying image sequence. GRU uses an extra memory state that depends on the input at the current time for the update. As a whole, GRU uses two sigmoid gates while RMU uses one sigmoid gate.

\vspace{+0.1em}\noindent\textbf{Hidden State Initialization.} Instead of initializing the first hidden state $h_{0}$ with zero, $\mathcal{F}$ resulting from the top level of the encoder is converted to $h_{0}$ as follows:
\begin{equation}
h_{0} = \text{tanh}(\text{convs}(\mathcal{F})).
\end{equation}

\vspace{+0.1em}\noindent\textbf{Depth Inference.} Depth map $D_{t}$ is inferred from the last hidden state. To prevent numerical issues during backpropagation, $D_{t}$ is bounded by $[D_{min},D_{max}]$ as follows:
\begin{subequations}
\begin{align} 
\hat{D}_{t} &= \sigma\left(\text{convs}\left(h_{k}\right)\right), \label{eq:7a} \\
D_{t} &= D_{min}(1-\hat{D}_{t}) + D_{max} \hat{D}_{t}. \label{eq:7b}
\end{align} 
\end{subequations}

\subsection{Residual Upsampling}\label{sec:residual upsampling}

Upsampling decoder feature is required when passing from a low-resolution to a high-resolution level in top-down approach~\cite{Godard19,Zhou17}. A feature map $x$ is upsampled to $x'$ by a upsampling function $f$ (such as deconvolution\cite{Zeiler11} or subpixel convolution~\cite{Shi16}). The process can be represented by
\begin{equation}
x' = \theta\left(f(x;\mathtt{W})\right),\label{eq:normal upsampling}
\end{equation}
where ``$\theta$'' denotes an activation function. Since a featuremap like a colour image consists of different spectral components, a single filter $\mathtt{W}$ is not universal enough to perform well on all regions. It is desired to use different upsampling filters on different regions (flat region: averaging filter, edge region: high-pass filter). To this end, a generic upsampling layer that uses multiple filters ${\mathtt{W}_{i}}$ is proposed as follows:
\begin{equation}
x' = \theta\biggl(\sum_{i}f_{i}(x;\mathtt{W}_{i})\biggr).
\end{equation}
Particularly, each upsampling operator $f_{i}$ is band-limited to some spectral components. The individual upsampled feature maps are summed before applying the activation. To compromise between accuracy and speed, RM-Depth is limited to use two kinds of upsampling operators, namely low-frequency $f_{l}$ and high-frequency $f_{h}$ ones, as follows:
\begin{equation}
x' = \theta\bigl(f_{l}(\text{conv}_{1\times1}(x)) + f_{h}(x;\mathtt{W}_{h})\bigr),
\end{equation}
where a 1$\times$1 convolution is used to squeeze $x$ for matching the channel dimension of $f_{h}(\cdot)$. A bilinear upsampling is chosen as $f_{l}$. Besides the 1$\times$1 convolution, there is no
additional increase in model parameters or computational overhead in comparison to Eq.~\eqref{eq:normal upsampling}.

\subsection{Object Motion}\label{sec:object motion}

Unsupervised learning of depth relies on novel view synthesis as presented in Sec.~\ref{sec:preliminaries}. Prior works tend to jointly recover depth and camera motion but leaving out motions of moving objects~\cite{Godard19,Guizilini20}. As a result, the visual displacement that is computed by Eq.~\eqref{eq:novel view synthesis} is just a component of full flow (so-called rigid flow) inferred by the camera motion. The novel view is not correctly synthesized and in turn affects the unsupervised training. Artifacts often exist in moving objects when the object motion is not taken into consideration (see Fig.~\ref{fig:depth predictions on Cityscapes} in Sec.~\ref{sec:results}). To resolve this issue, both the camera and object motions are necessarily recovered. Since it is rare to have objects spinning on their owns with large magnitudes in street-view scenes, it can be assumed that the rotational motion of moving objects is nearly zero. An overview of the proposed motion network is shown in Fig.~\ref{fig:motion network}. More details are presented below.

\vspace{+0.1em}\noindent\textbf{Warping-Based Motion Field Inference.} Motions of moving objects are estimated in form of a motion field $T_{obj}: \Omega \rightarrow \mathbb{R}^{3}$ in a coarse-to-fine framework as shown in Fig.~\ref{fig:motion network}. The motion field $T_{obj}$ is combined with camera motion tcam to form a complete motion field. Source images $\{I_{s}\}$ are warped towards the target image $I_{t}$ in accordance with the full flow ${\bf u}_{full} = {\bf x}_{s} - {\bf x}_{t}$, where ${\bf x}_{s}$ is computed by Eq.~\eqref{eq:novel view synthesis}. For the initialization, $\{I_{s}\}$ are warped towards $I_{t}$ in accordance with the rigid flow by setting $T_{obj} = {\bf 0}$. The warped source images $\{I_{s\rightarrow t}\}$ together with the target image $I_{t}$ are fed into the motion encoder to generate a new set of multi-scale encoder features $\{\mathcal{F}(I_{t},I_{s \rightarrow t})\}$. The encoder features are more aligned to $I_{t}$ since $\{I_{s}\}$ have been warped towards $I_{t}$. This in turn makes the generation of motion field easier as inspired by the feature warping proposed in LiteFlowNet series~\cite{Hui18,Hui20,Hui21}. The object motion decoder refines the previous estimate $T_{obj,l+1}$ by augmenting with the encoder feature at the same scale as follows:
\begin{equation}
T_{obj,l} = \text{convs}([T_{obj,l+1}^{\uparrow2}, \mathcal{F}_{l}(I_{t},I_{s \rightarrow t})]) + T_{obj,l+1}^{\uparrow 2},
\end{equation}
where ``$\text{convs}$'' represents several convolution layers and $(\cdot)^{\uparrow 2}$ denotes an upsampling operator by a factor of 2. Particularly, the encoder features are warped for the motion refinement. This is different from prior works~\cite{Gordon19,Li20} that use fixed encoder features.

\begin{figure}[t]
\centering
\includegraphics[width=\linewidth]{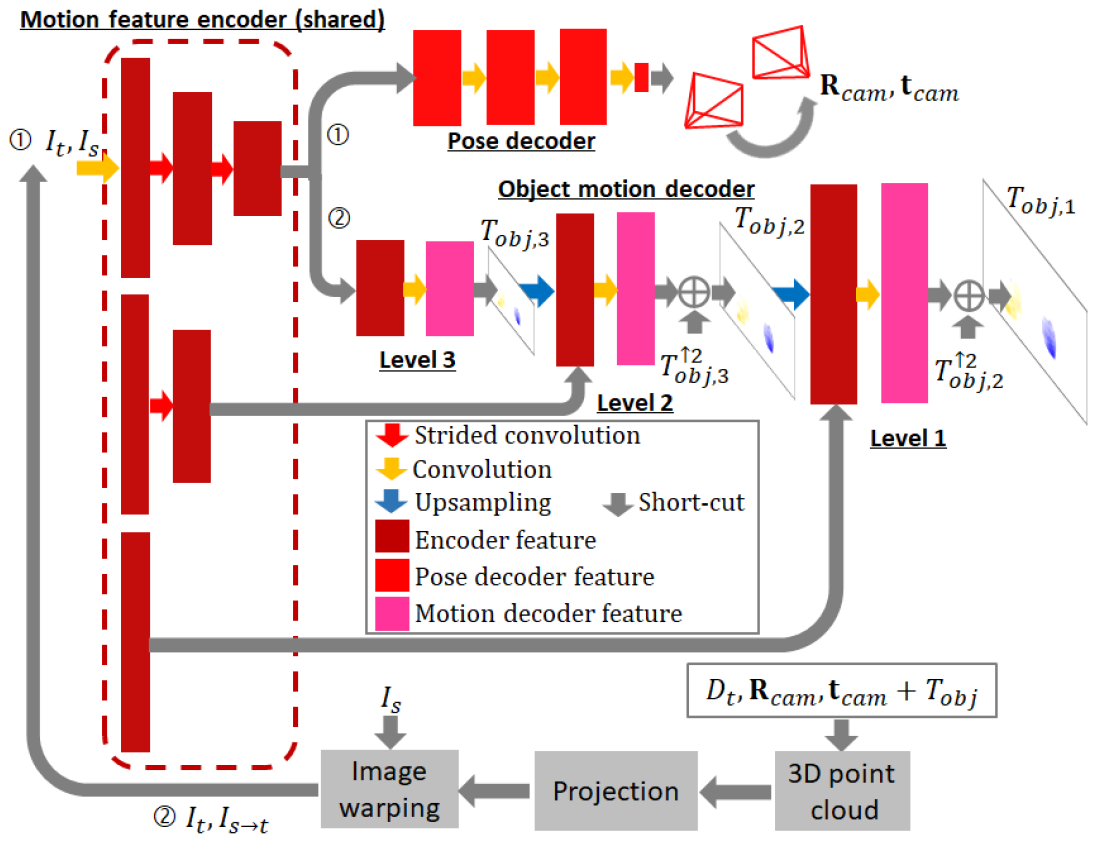}
\caption{The architecture of the proposed motion network. The encoder is shared by the pose and object motion decoders. Object motion field $T_{obj}$ is refined in a multi-scale framework by feedbackwarding the previous estimate to the encoder through novel view synthesis (see Sec.~\ref{sec:preliminaries}).}\label{fig:motion network}
\end{figure}

\vspace{+0.1em}\noindent\textbf{Outlier-Aware Regularization Loss.} Motion field is generally sparse since moving objects do not fully occupy a scene, \textit{i.e.} $T_{obj}({\bf x}) = {\bf 0}$ when an image position {\bf x} is not affected by non-rigid motion. This observation can impose a constraint on the unsupervised training and in turn improves the depth accuracy. A motion mask $M$ is constructed by comparing full flow ${\bf u}_{full}$ (computed by Eq.~\eqref{eq:novel view synthesis} using depth, camera and object motions) against rigid flow urig (using only depth and camera motion). If there are no moving objects in the scene other than the moving camera itself, then ${\bf u}_{full} = {\bf u}_{rig}$. Otherwise, ${\bf u}_{full} \neq {\bf u}_{rig}$. This motivation allows us to segment image locations affected by non-rigid motions using the following condition:
\begin{equation}
M({\bf x}) = [||{\bf u}_{full} - {\bf u}_{rig}||_{2} < \alpha],
\end{equation}
where $[\cdot]$ is the Iverson bracket. A thresholding approach is used to suppress outliers by setting  $\alpha = 0.5$. When an image position ${\bf x}$ is affected by non-rigid motions, $M(x) = 0$. Otherwise, $M({\bf x}) = 1$. With the motion mask, an outlieraware regularization loss $L_{reg}$ on the motion field is proposed as follows:
\begin{equation}
L_{reg}(T_{obj}) = \sum_{x \in \Omega} g(M \cdot T_{obj}),
\end{equation}
where $g(\cdot)$ is chosen to be the sparsity function~\cite{Li20} as it encourages more sparsity than $L_{1}$ norm. $L_{reg}$ helps the motion network to properly learn $T_{obj}$ by suppressing the growth of undesired object motion in rigid regions.

\section{Experiments}\label{sec:experiments}
\subsection{Implementation Details}

\noindent\textbf{Network Architecture.} The overviews of depth and motion networks can be referred to Figs.~\ref{fig:different depth models} and~\ref{fig:motion network}, respectively. A modified 6-level ResNet18~\cite{He16} that contains an additional convolution layer at the bottom level and excludes the classification head is adopted as the encoders. Particularly, the top two levels are not used in the depth encoder. For the motion network, the pose decoder is adopted from~\cite{Godard19}. The object motion decoder\footnote{The first bottom level of the object motion decoder is modified compared with the CVPR version~\cite{Hui22}} uses 9 and 2 RMUs in level 4 and the remained levels, respectively. RMUs are not shared across different levels in order to maximize filter diversity for different scales. 

\vspace{+0.1em}\noindent\textbf{Training Details.} The whole system is implemented in TensorFlow~\cite{Abadi15}. Same augmentations are performed on the training data as~\cite{Godard19}, namely 50\% horizontal flips, random brightness, contrast, saturation, and hue jitter. Following~\cite{Zhou17}, the length of each image sequence is fixed to 3 frames. The central frame is treated as the target view. The depth and motion networks are jointly trained using Adam~\cite{Kingma15} with a batch size varying from 16 to 40 on multiple GPUs. To address the stationary pixels and the occlusion problem, the auto-masking and the per-pixel minimum reprojection loss~\cite{Godard19} are adopted. Depth map and motion field are regularized by an edge-aware smoothness loss~\cite{Godard19} while the proposed outlier-aware regularization loss is further imposed on the object motion field. The self-supervision~\cite{Stone21} is also adopted but no cropping is applied. Some parts of RM-Depth require pre-training\footnote{This is different from the CVPR version~\cite{Hui22}.}. After that, the overall network is trained for 20 epochs. A learning rate of 1e-4 for the first 10 epochs and reduce the learning rate to 1e-5 for the remained epochs. All the encoders have been pre-trained on ImageNet~\cite{Russakovsky15}. 

\vspace{+0.1em}\noindent\textbf{Dataset.} The system is trained and validated on KITTI~\cite{Geiger13} and Cityscapes~\cite{Cordts16}. The image resolution is set to 640$\times$192. For KITTI, the data split of Eigen~\textit{et al.}~\cite{Eigen14} that excludes all the evaluation frames is used as the training set. For an evaluation, static frames are excluded so that it is comparable to Zhou~\textit{et al.}~\cite{Zhou17}. For Cityscapes, the standard training split is used and no static frames are neglected. The cropping scheme ``A'' defined in~\cite{Watson21} is used for the evaluation.

\subsection{Results}\label{sec:results}

\begin{table*}[t]
\centering
\caption{Monocular depth results on the KITTI dataset (K) by the testing split of Eigen \textit{et al.}~\cite{Eigen15} and the testing split of Cityscapes dataset (CS). Models that require explicit semantic data are highlighted. The best in each category is in bold and the second best is underlined.}
\scalebox{0.72}{
\begin{tabular}{l|c|c|c|cccc|ccc} 
 \thickhline
 Method & Semantics & Training & Testing & \multicolumn{4}{c|}{Error (lower is better)} & \multicolumn{3}{c}{Accuracy (higher is better)}\\ 
   & & datset & datset & AbsRel & SqRel & RMS & RMSlog & $\delta<1.25$ & $\delta<1.25^{2}$ & $\delta<1.25^{3}$\\ 
 \hline
 Zhou \textit{et al.}~\cite{Zhou17}                          & & K & K & 0.208 & 1.768 & 6.856 & 0.283 & 0.678 & 0.885 & 0.957\\ 
 GeoNet~\cite{Yin18}                                             & & K & K & 0.164 & 1.303 & 6.090 & 0.247 & 0.765 & 0.919 & 0.968\\
 Mahjourian \textit{et al.}~\cite{Mahjourian18}       & & K & K & 0.163 & 1.240 & 6.220 & 0.250 & 0.762 & 0.916 & 0.968\\ 
 GeoNet~\cite{Yin18}                                             & & K & K & 0.155 & 1.296 & 5.857 & 0.233 & 0.793 & 0.931 & 0.973\\ 
 DDVO~\cite{Wang18}                                          & & K & K & 0.151 & 1.257 & 5.583 & 0.228 & 0.810 & 0.936 & 0.974\\
 Li \textit{et al.}~\cite{Li19}                                     & & K & K & 0.150 & 1.127 & 5.564 & 0.229 & 0.823 & 0.936 & 0.974\\ 
 DF-Net~\cite{Zou18}                                            & & K & K & 0.150 & 1.124 & 5.507 & 0.223 & 0.806 & 0.933 & 0.973\\  
 Pilzer~\textit{et al.}~\cite{Andrea19}                    & & K & K & 0.142 & 1.231 & 5.785 & 0.239 & 0.795 & 0.924 & 0.968\\
 EPC++~\cite{Luo20}                                            & & K & K & 0.141 & 1.029 & 5.350 & 0.216 & 0.816 & 0.941 & 0.976\\ 
 Struct2Depth~\cite{Casser19}             & $\bullet$ & K & K & 0.141 & 1.026 & 5.291 & 0.215 & 0.816 & 0.945 & 0.979\\ 
 CC~\cite{Ranjan19}                                              & & K & K & 0.140 & 1.070 & 5.326 & 0.217 & 0.826 & 0.941 & 0.975\\ 
 Bian~\textit{et al.}~\cite{Bian19}                           & & K & K & 0.137 & 1.089 & 5.439 & 0.217 & 0.830 & 0.942 & 0.975\\
 GLNet~\cite{Chen19}                                           & & K & K & 0.135 & 1.070 & 5.230 & 0.210 & 0.841 & 0.948 & 0.980\\
 Li~\textit{et al.}\cite{Li20}                      & $\bullet$ & K & K & 0.130 & 0.950 & 5.138 & 0.209 & 0.843 & 0.948 & 0.978\\
 Gordon~\textit{et al.}~\cite{Gordon19}& $\bullet$ & K & K & 0.128 & 0.959 & 5.230 & 0.212 & 0.845 & 0.947 & 0.976\\
 Distilled Semantics~\cite{Tosi20}        & $\bullet$ & K & K & 0.126 & 0.835 & 4.937 & 0.199 & 0.844 & 0.953 & 0.982\\
 Monodepth2~\cite{Godard19}                             & & K & K & 0.115 & 0.882 & 4.701 & 0.190 & 0.879 & \underline{0.961} & 0.982\\
 PackNet~\cite{Guizilini20}                                    & & K & K & \underline{0.111} & 0.785 & \underline{4.601} & 0.189 & 0.878 & 0.960 & 0.982\\
 PackNet~\cite{Guizilini20} (with velocity weak supervision)  & & K & K & \underline{0.111} & 0.829 & 4.788 & 0.199 & 0.864 & 0.954 & 0.980\\
 Johnston~\textit{et al.}~\cite{Johnston20}          & & K & K & \underline{0.111} & 0.941 & 4.817 & 0.189 & \textbf{0.885} & \underline{0.961} & 0.981\\
 Monodepth2-Boot+Self~\cite{Poggi20}             & & K & K & \underline{0.111} & 0.826 & 4.667 & \underline{0.184} & 0.880 & \underline{0.961} & \underline{0.983}\\
 Monodepth2-Boot+Log~\cite{Poggi20}             & & K & K & 0.117 & 0.900 & 4.838 & 0.192 & 0.873 & 0.958 & 0.981\\
 Lee~\textit{et al.}~\cite{Lee21}            & $\bullet$ & K & K & 0.112 & \underline{0.777} & 4.772 & 0.191 & 0.872 & 0.959 & 0.982\\
 Gao~\textit{et al.}~\cite{Gao20}                          & & K & K & 0.112 & 0.866 & 4.693 & 0.189 & 0.881 & \underline{0.961} & 0.981\\
 \textbf{RM-Depth}                                                & & K & K & \textbf{0.107} & \textbf{0.687} & \textbf{4.476} & \textbf{0.181} & \underline{0.883} & \textbf{0.964} & \textbf{0.984}\\
 \hline
 Zhou \textit{et al.}~\cite{Zhou17}                          & & CS + K & K & 0.198 & 1.836 & 6.565 & 0.275 & 0.718 & 0.901 & 0.960\\
 Mahjourian \textit{et al.}~\cite{Mahjourian18}       & & CS + K & K & 0.159 & 1.231 & 5.912 & 0.243 & 0.784 & 0.923 & 0.970\\ 
 GeoNet~\cite{Yin18}                                             & & CS + K & K & 0.153 & 1.328 & 5.737 & 0.232 & 0.802 & 0.934 & 0.972\\
 DDVO~\cite{Wang18}                                          & & CS + K & K & 0.148 & 1.187 & 5.496 & 0.226 & 0.812 & 0.938 & 0.975\\
 DF-Net~\cite{Zou18}                                            & & CS + K & K & 0.146 & 1.182 & 5.215 & 0.213 & 0.818 & 0.943 & 0.978\\
 PackNet~\cite{Guizilini20}                                   & & CS + K & K & \underline{0.108} & \underline{0.727} & \underline{4.426} & \underline{0.184} & \underline{0.885} & \underline{0.963} & \textbf{0.984}\\
 PackNet~\cite{Guizilini20} (with velocity weak supervision) & & CS + K & K & \underline{0.108} & 0.803 & 4.642 & 0.195 & 0.875 & 0.958 & \underline{0.980}\\
 \textbf{RM-Depth}                                                & & CS + K & K & \textbf{0.105} & \textbf{0.675} & \textbf{4.368} & \textbf{0.178} & \textbf{0.889} & \textbf{0.965} & \textbf{0.984}\\  
 \hline
 Struct2Depth~\cite{Casser19}            & $\bullet$ & CS & CS & 0.145 & 1.737 & 7.280 & 0.205 & 0.813 & 0.942 & 0.976\\
 GLNet~\cite{Chen19} (with online refinement)  & & CS & CS & 0.129 & \underline{1.044} & \textbf{5.361} & 0.212 & 0.843 & 0.938 & 0.976\\
 Gordon~\textit{et al.}~\cite{Gordon19}& $\bullet$& CS & CS & 0.127 & 1.330 & 6.960 & 0.195 & 0.830 & 0.947 & 0.981\\
 Li~\textit{et al.}~\cite{Li20}                                  & & CS & CS & 0.119 & 1.290 & 6.980 & 0.190 & 0.846 & 0.952 & 0.982\\
 Lee~\textit{et al.}~\cite{Lee21}            & $\bullet$ & CS & CS & \underline{0.111} & 1.158 & 6.437 & \underline{0.182} & \underline{0.868} & \underline{0.961} & \underline{0.983}\\
 \textbf{RM-Depth}                                                & & CS & CS & \textbf{0.090} & \textbf{0.825} & \underline{5.503} & \textbf{0.143} & \textbf{0.913} & \textbf{0.980} & \textbf{0.993}\\
 
 \thickhline
\end{tabular}}
\label{tab:results on KITTI and Cityscapes}
\end{table*}

RM-Depth is compared against prior state-of-the-art methods such that they are also trained on monocular image sequences and perform single-image depth inference without using online refinement unless otherwise specified. Depth map is capped to 80m~\cite{Godard17} and is normalized using median scaling~\cite{Zhou17}. Other experimental results (related to generalization on unseen dataset, visual odometry, and more) are available in the supplementary material~\cite{Hui22sup}.

\begin{figure*}[t]
\captionsetup[subfigure]{labelformat=empty}
\centering
\subfloat[]{\includegraphics[height=1.05cm]{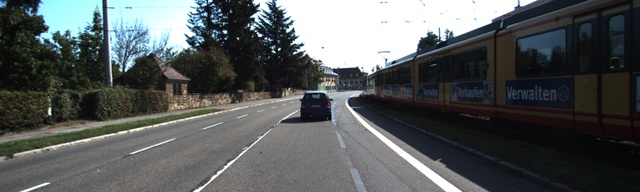}}
\subfloat[]{\includegraphics[height=1.05cm]{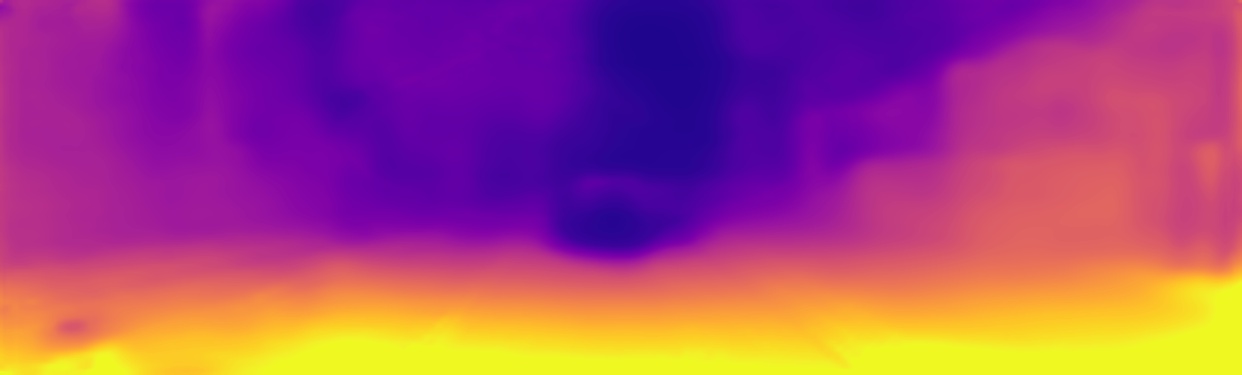}}
\subfloat[]{\includegraphics[height=1.05cm]{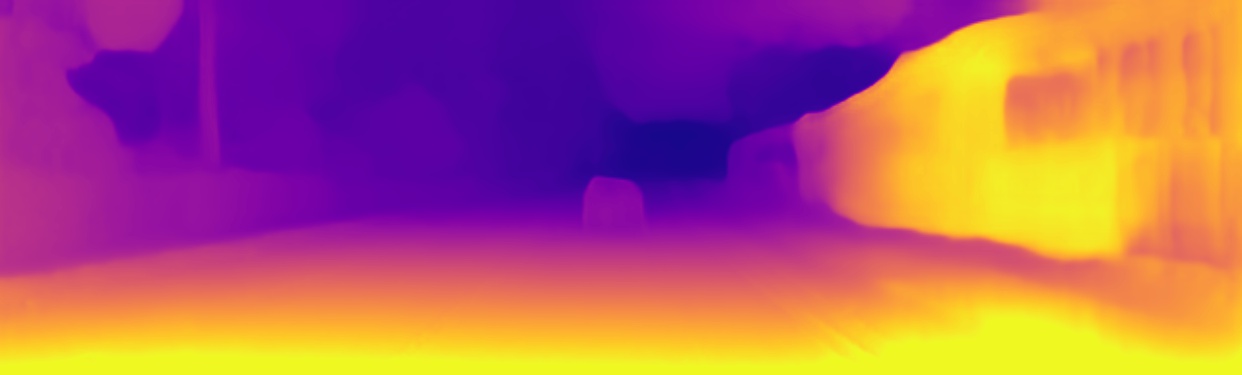}}
\subfloat[]{\includegraphics[height=1.05cm]{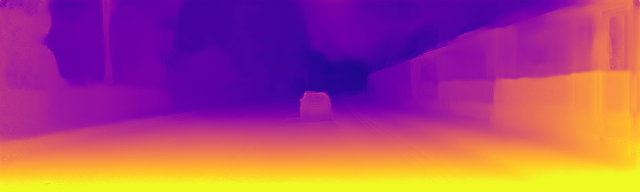}}
\subfloat[]{\includegraphics[height=1.05cm]{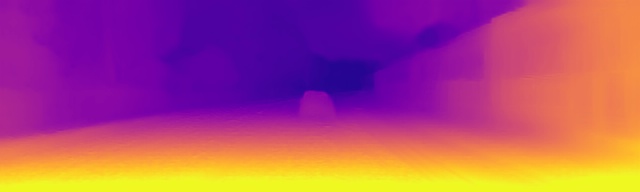}}\\ \vspace{-1.2em}
\subfloat[]{\includegraphics[height=1.05cm]{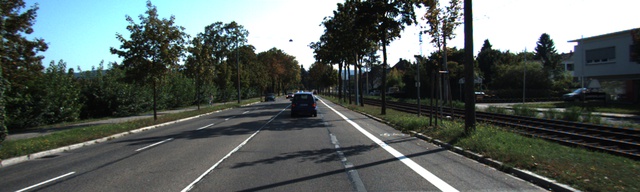}}
\subfloat[]{\includegraphics[height=1.05cm]{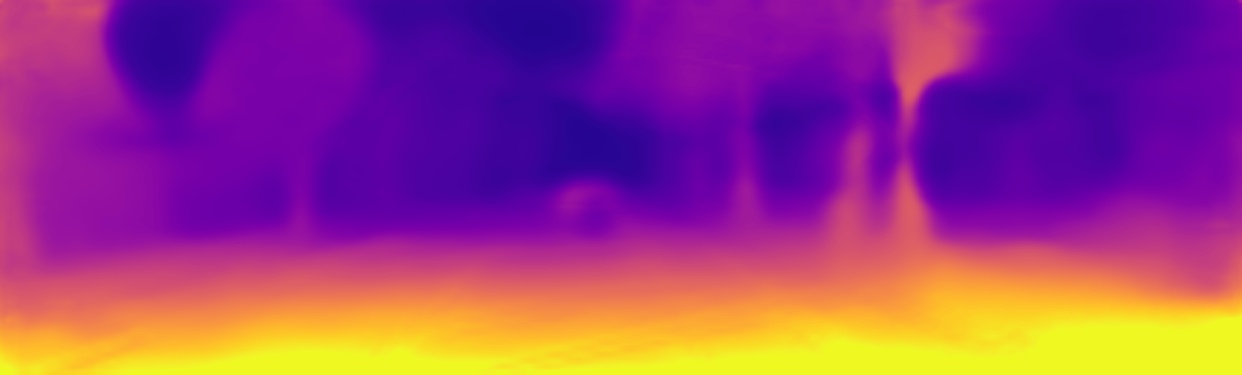}}
\subfloat[]{\includegraphics[height=1.05cm]{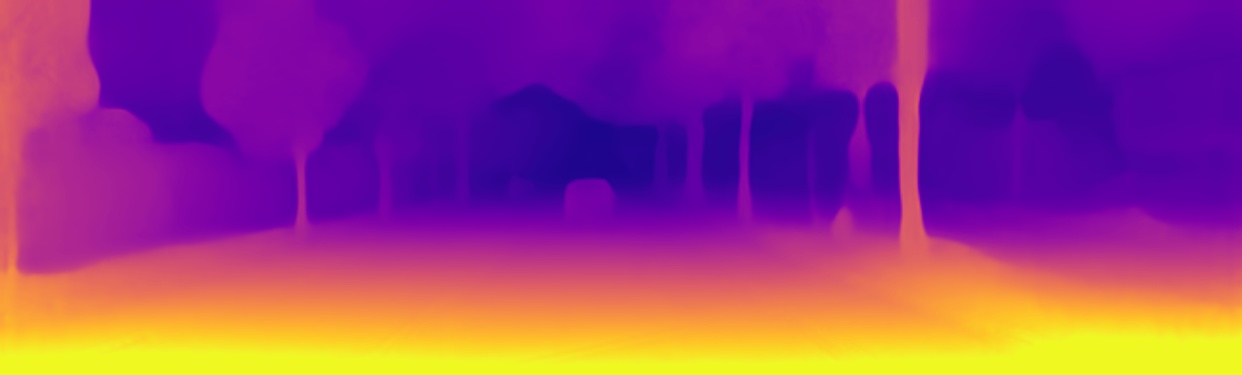}}
\subfloat[]{\includegraphics[height=1.05cm]{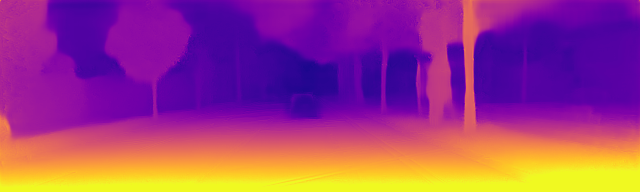}}
\subfloat[]{\includegraphics[height=1.05cm]{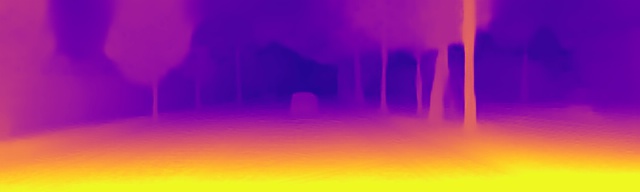}}\\  \vspace{-1.2em}
\subfloat[RGB image]{\includegraphics[height=1.05cm]{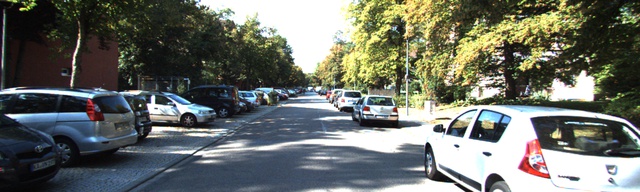}}
\subfloat[GeoNet~\cite{Yin18}]{\includegraphics[height=1.05cm]{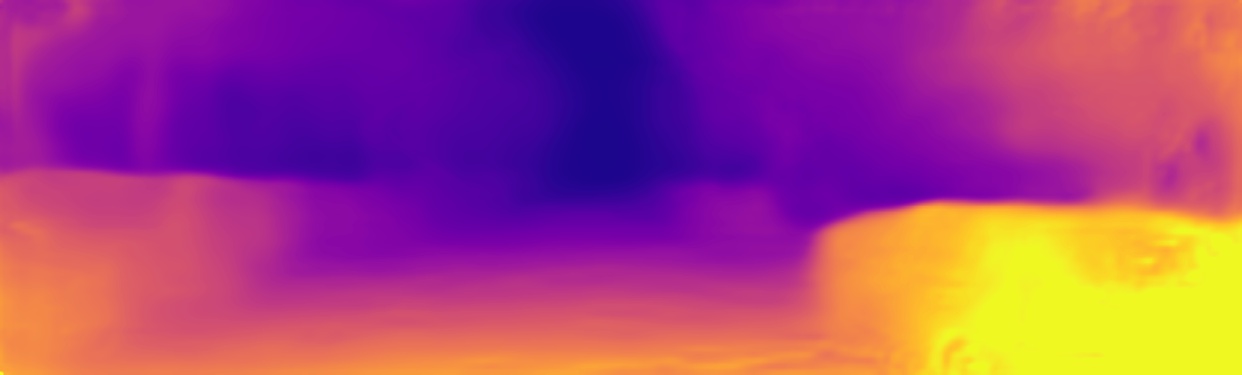}}
\subfloat[Monodepth2~\cite{Godard19}]{\includegraphics[height=1.05cm]{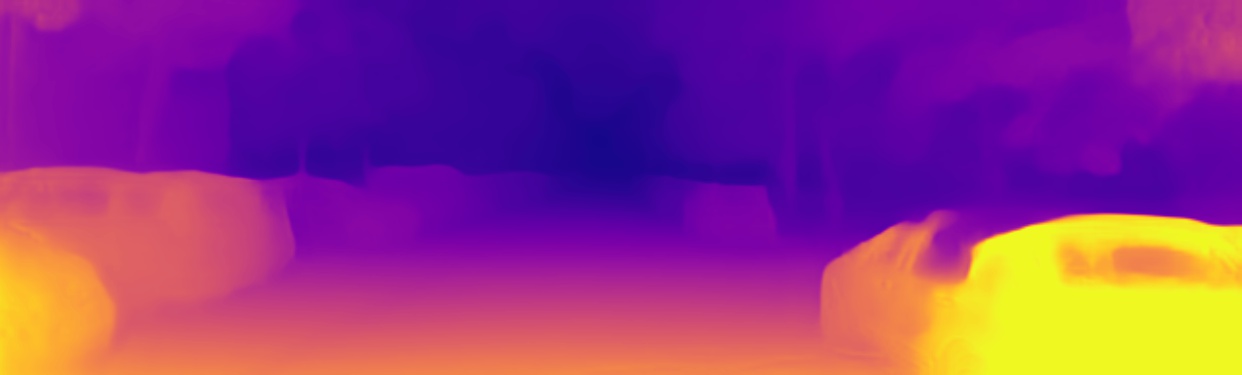}}
\subfloat[PackNet~\cite{Guizilini20}]{\includegraphics[height=1.05cm]{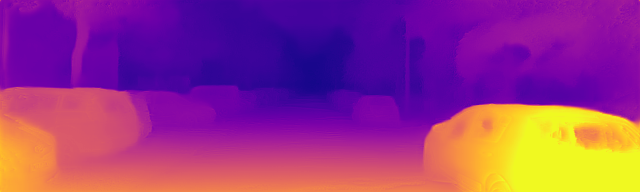}}
\subfloat[RM-Depth]{\includegraphics[height=1.05cm]{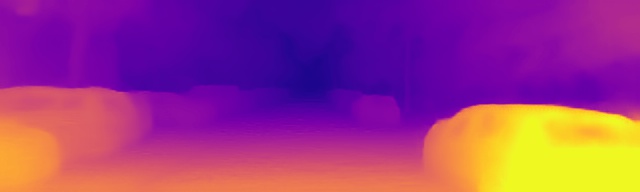}}
\caption{Examples of depth map predictions on KITTI.}\label{fig:depth predictions on KITTI}
\end{figure*}

\vspace{+0.1em}\noindent\textbf{Depth (KITTI).} As shown in the upper half of Table~\ref{tab:results on KITTI and Cityscapes}, RM-Depth outperforms the compared methods. Examples of estimations are provided in Fig.~\ref{fig:depth predictions on KITTI}. It can be observed that RM-Depth is superior in recovering thin structures and moving objects than GeoNet~\cite{Yin18}. Monodepth2~\cite{Godard19} cannot correctly predict depth values on objects with reflective surface (the on-road train in the first example and the white car in the third example) while RM-Depth is free of such defects. PackNet~\cite{Guizilini20} and RM-Depth recover depth maps with sharp discontinuities. However the moving car in the second example is not correctly estimated by PackNet.

\begin{figure*}[t]
\captionsetup[subfigure]{labelformat=empty}
\centering
\subfloat[]{\includegraphics[height=1.05cm]{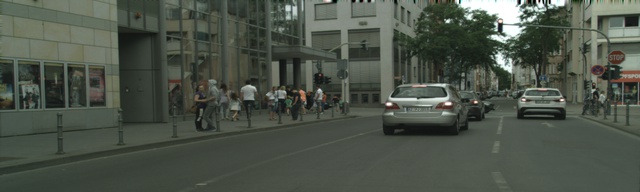}}
\subfloat[]{\includegraphics[height=1.05cm]{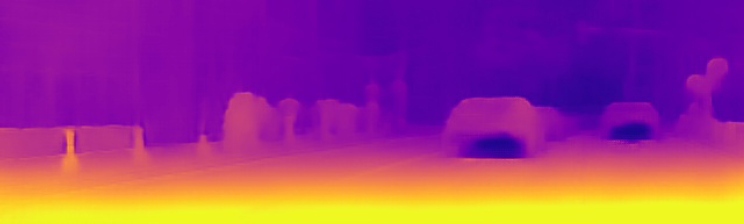}}
\subfloat[]{\includegraphics[height=1.05cm]{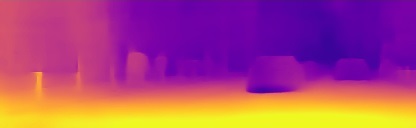}}
\subfloat[]{\includegraphics[height=1.05cm]{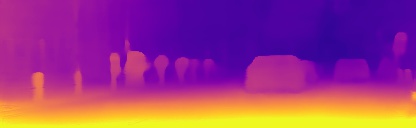}}
\subfloat[]{\includegraphics[height=1.05cm]{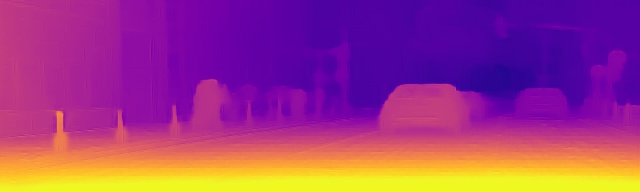}}\\ \vspace{-1.2em}
\subfloat[RGB image]{\includegraphics[height=1.05cm]{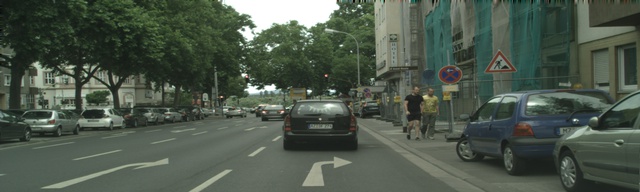}}
\subfloat[without object motion]{\includegraphics[height=1.05cm]{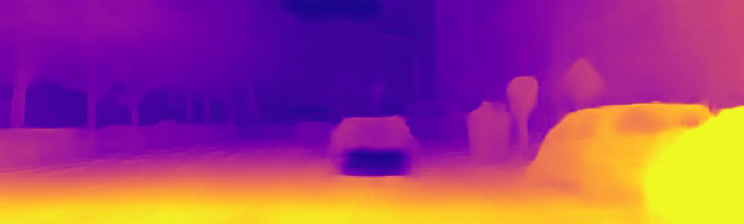}}
\subfloat[Struct2Depth~\cite{Casser19}]{\includegraphics[height=1.05cm]{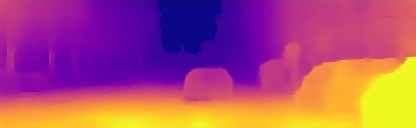}}
\subfloat[Gordon~\textit{et al.}~\cite{Gordon19}]{\includegraphics[height=1.05cm]{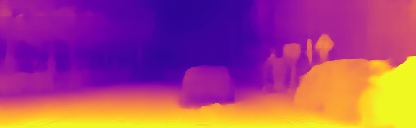}}
\subfloat[RM-Depth]{\includegraphics[height=1.05cm]{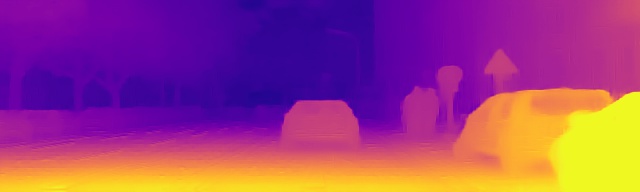}}\\
\caption{Examples of depth map predictions on Cityscapes.}\label{fig:depth predictions on Cityscapes}
\end{figure*}

\vspace{+0.1em}\noindent\textbf{Depth (Cityscapes).} This dataset is more challenging as it involves more moving objects than KITTI. Only a few works report the evaluation results on Cityscapes. The bottom half of Table~\ref{tab:results on KITTI and Cityscapes} summarizes the results. Despite RM-Depth does not use segmentation labels, it outperforms the prior works. Visual comparison is provided in Fig.~\ref{fig:depth predictions on Cityscapes}. When object motion is neglected, holes (\textit{i.e.} depth values tend to the maximum) often appear on moving objects.

\begin{figure*}[t]
\captionsetup[subfigure]{labelformat=empty}
\centering
\subfloat[]{\includegraphics[height=1.05cm]{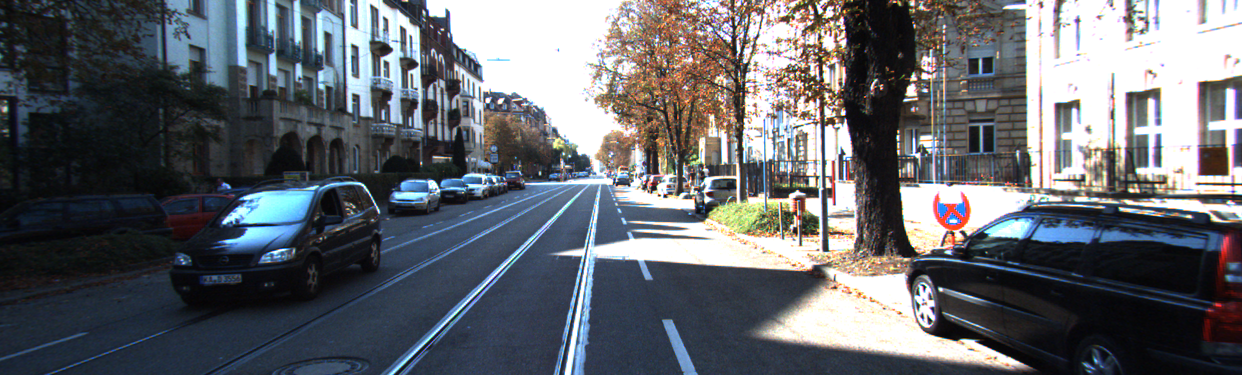}} 
\subfloat[]{\includegraphics[height=1.05cm]{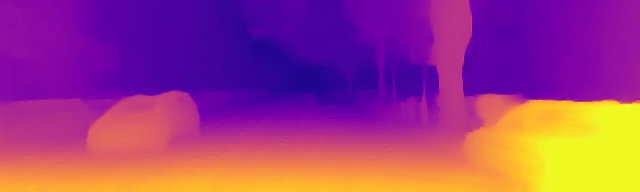}}
\subfloat[]{\includegraphics[height=1.05cm]{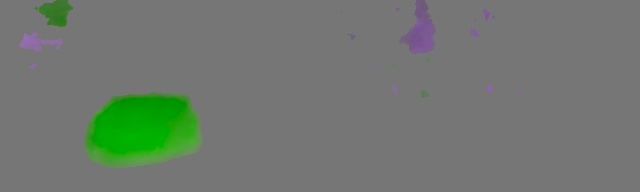}}
\subfloat[]{\includegraphics[height=1.05cm]{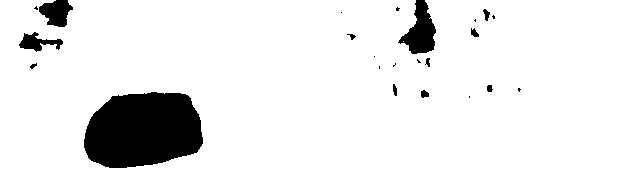}}
\subfloat[]{\includegraphics[height=1.05cm]{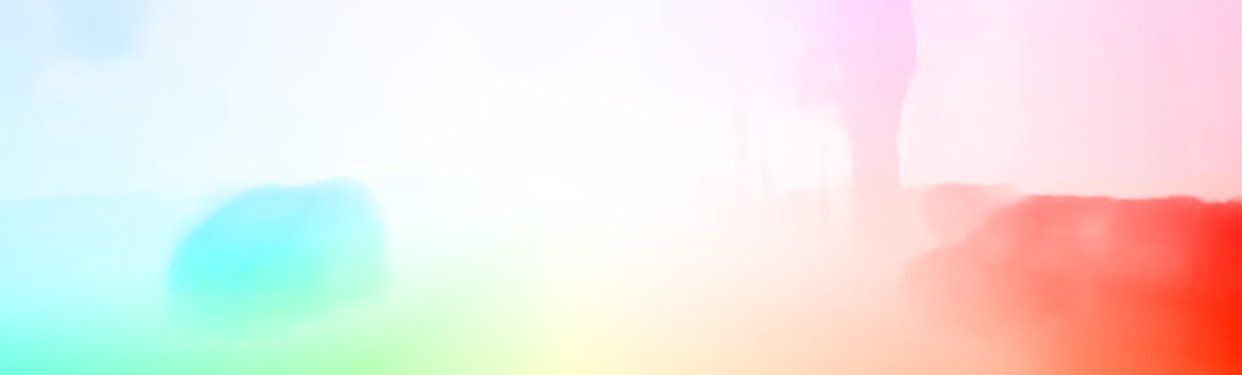}}\\ \vspace{-1.2em}

\subfloat[]{\includegraphics[height=1.05cm]{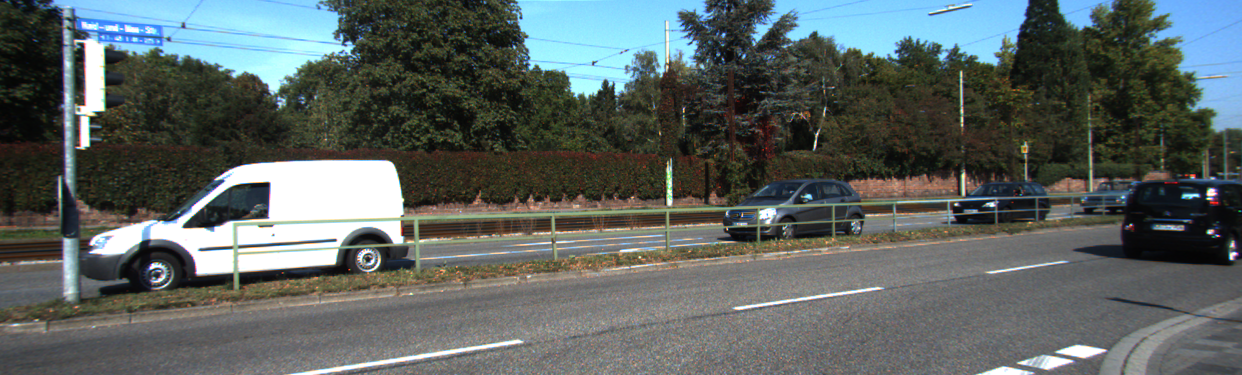}} 
\subfloat[]{\includegraphics[height=1.05cm]{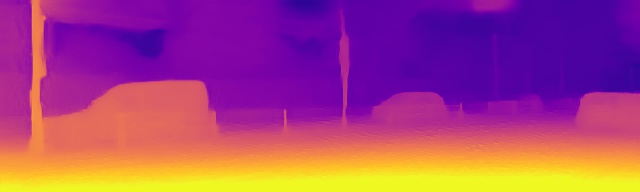}}
\subfloat[]{\includegraphics[height=1.05cm]{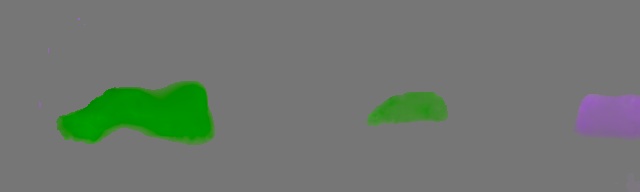}}
\subfloat[]{\includegraphics[height=1.05cm]{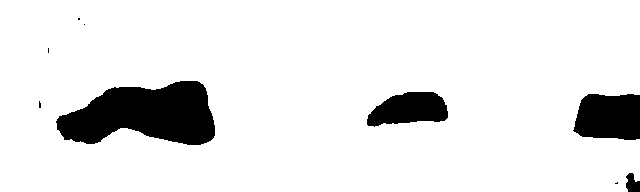}}
\subfloat[]{\includegraphics[height=1.05cm]{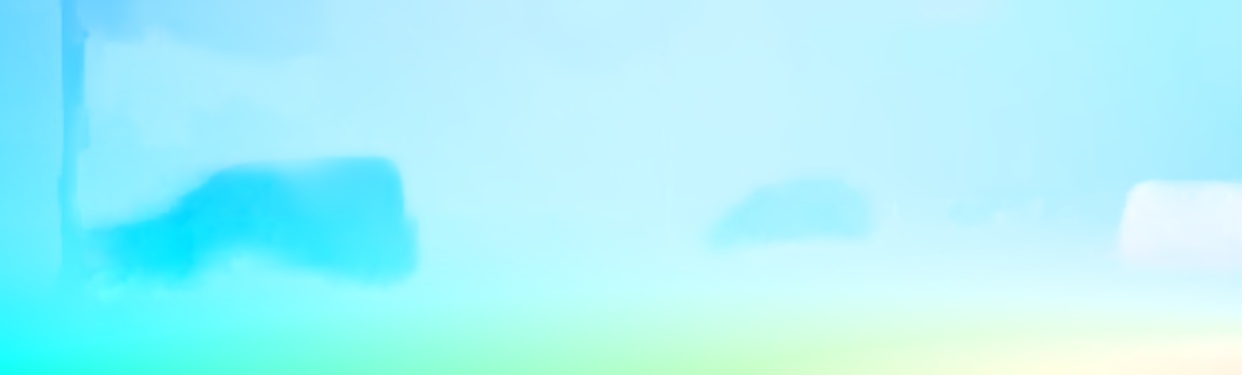}}\\ \vspace{-1.2em}

\subfloat[]{\includegraphics[height=1.05cm]{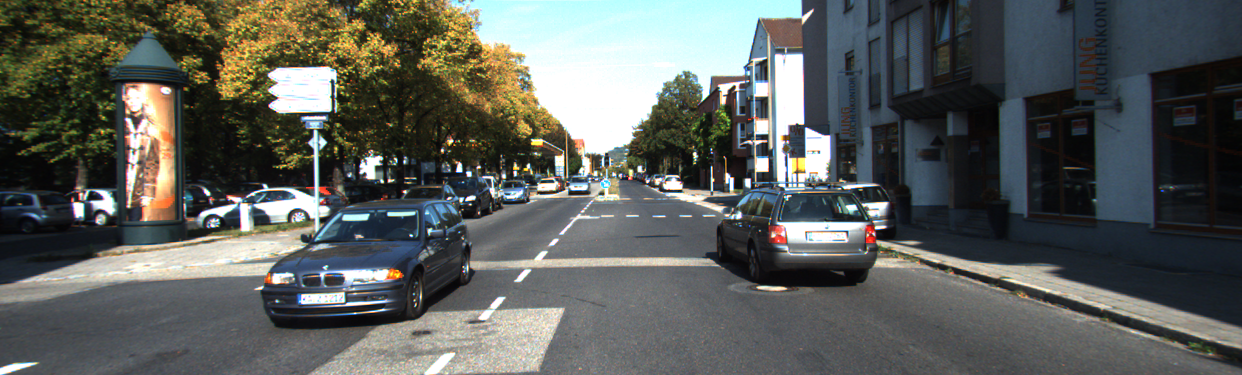}} 
\subfloat[]{\includegraphics[height=1.05cm]{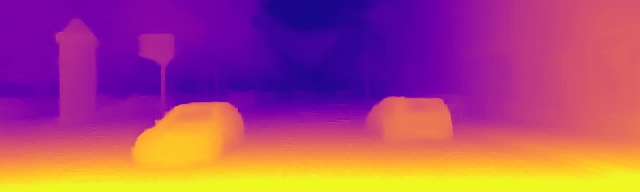}}
\subfloat[]{\includegraphics[height=1.05cm]{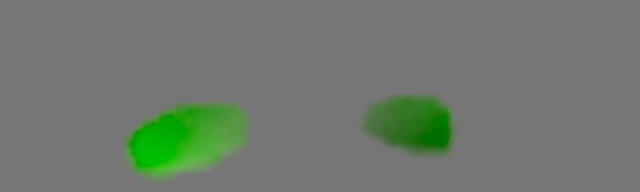}}
\subfloat[]{\includegraphics[height=1.05cm]{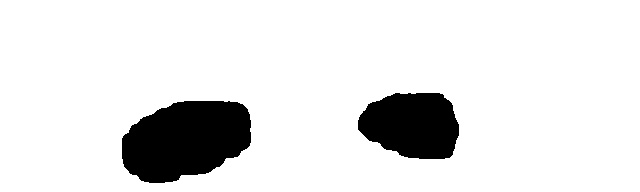}}
\subfloat[]{\includegraphics[height=1.05cm]{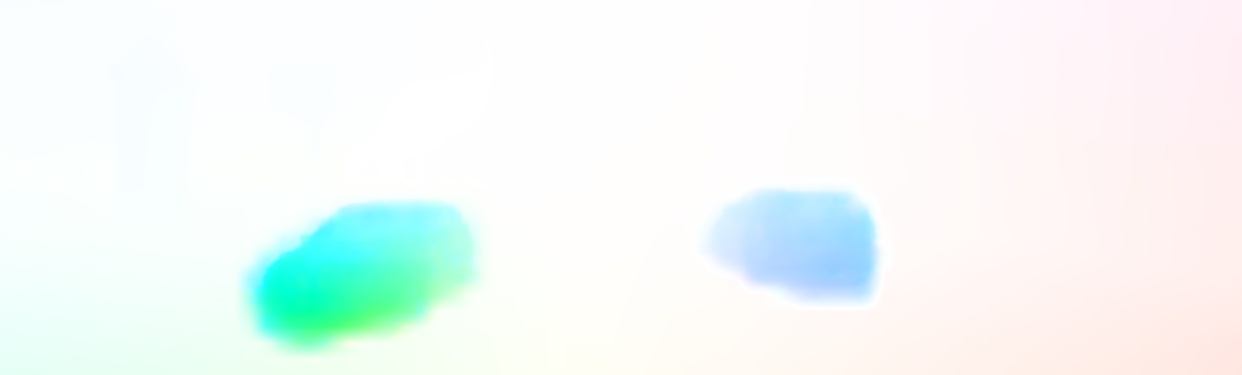}}\\ \vspace{-1.2em}

\subfloat[RGB image]{\includegraphics[height=1.05cm]{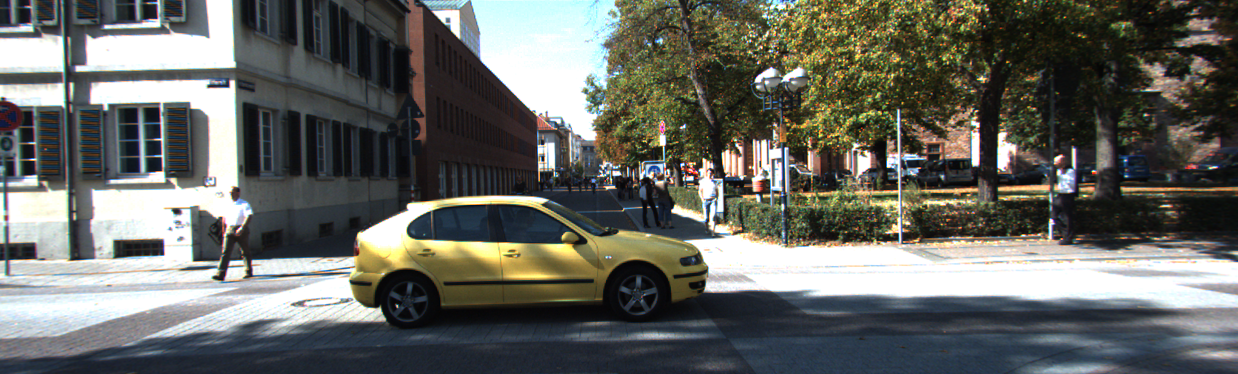}} 
\subfloat[Depth]{\includegraphics[height=1.05cm]{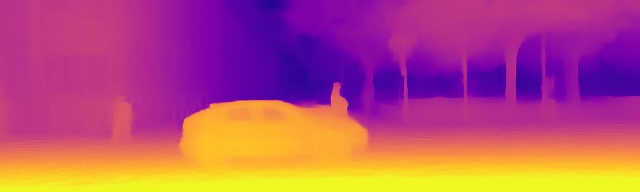}}
\subfloat[Motion field]{\includegraphics[height=1.05cm]{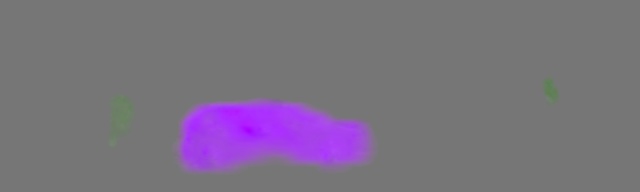}}
\subfloat[Segmentation mask]{\includegraphics[height=1.05cm]{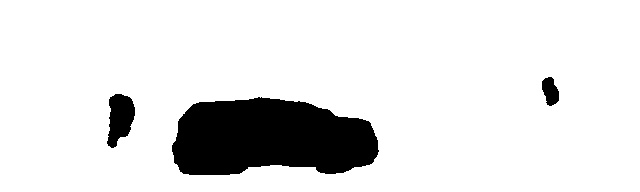}}
\subfloat[Optical flow]{\includegraphics[height=1.05cm]{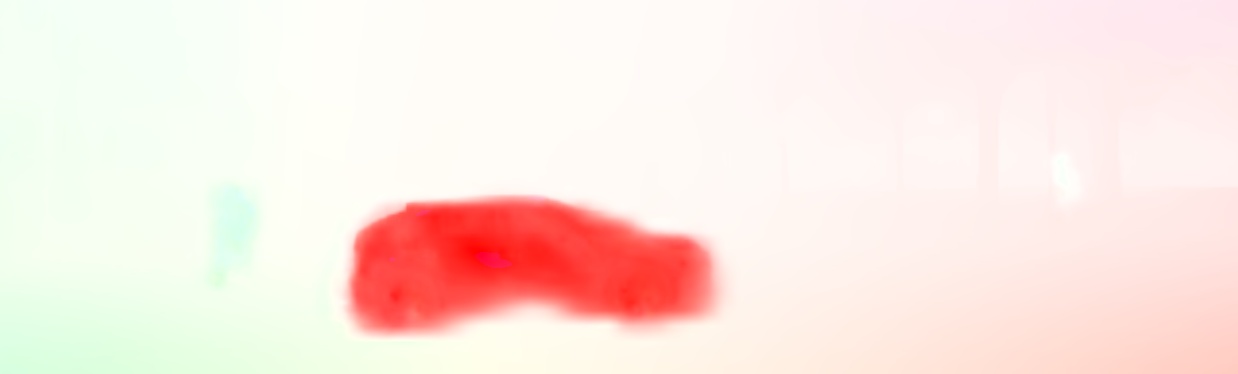}}

\caption{Examples of depth, object motion field, segmentation mask, and optical flow predictions on the KITTI 2015 dataset.}\label{fig:all results on KITTI}
\end{figure*}

\vspace{+0.1em}\noindent\textbf{Object Motion and Segmentation.} The protocol as~\cite{Ranjan19} is followed and the motion segmentation is evaluated on the KITTI 2015 dataset~\cite{Menze15}. The results are summarized in Table~\ref{tab:motion segmentation results on KITTI}. RM-Depth outperforms the compared methods including Distilled Semantics~\cite{Tosi20} while RM-Depth neither uses semantic labels for training nor semantic network. Fig.~\ref{fig:all results on KITTI} show examples of motion field and segmentation predictions.

\begin{table}[t]
\centering
\caption{Motion segmentation results on the KITTI 2015 dataset.}\label{tab:motion segmentation results on KITTI}
\vspace{-0.5em}
\scalebox{0.72}{
\begin{tabular}{l|c|ccc} 
\thickhline
 Model & Semantics & \multicolumn{3}{c}{Intersection over Union (IoU)} \\ 
            & & Overall & Static car & Moving car\\ 
 \hline
 EPC++~\cite{Luo20}                           &                 & 50.00             & -                      & - \\ 
 CC~\cite{Ranjan19}                             & $\bullet$ & 56.94             & 55.77              & 58.11\\
 DS~\cite{Tosi20}                                  & $\bullet$ & 62.66             & 58.42              & 66.89\\
 DS (semantic network)~\cite{Tosi20} & $\bullet$ & 63.98             & 64.16              & 63.79\\
 \textbf{RM-Depth}                                &                & \textbf{72.12} & \textbf{71.87} & \textbf{72.37}\\
 \thickhline
\end{tabular}}  
\end{table}

\vspace{+0.1em}\noindent\textbf{Optical Flow.} It is computed by Eq.~\eqref{eq:novel view synthesis} using depth, camera and object motions. As provided in Table~\ref{tab:optical flow results on KITTI}, AEE is improved when object motion is considered. The performance is reasonable since no stand-alone optical flow network is constructed. Examples of optical flow are shown in Fig.~\ref{fig:all results on KITTI}.

\begin{table}[t]
\centering
\caption{Optical flow results in terms of average end-point error on the KITTI 2015 dataset.}\label{tab:optical flow results on KITTI}
\vspace{-0.5em}
\scalebox{0.72}{
\begin{tabular}{l|c|cc} 
\thickhline
 Model & Explicit & All & F1\\ 
 & flow network & &\\ 
 \hline
 Distilled Semantics (ego-motion)~\cite{Tosi20} &                & 13.50             & 51.22\% \\ 
 Distilled Semantics~\cite{Tosi20}                       & $\bullet$ & 11.61 & \textbf{25.78}\% \\ 
 GeoNet (DirFlowNetS)~\cite{Yin18}                   & $\bullet$ & 12.21              & - \\   
 GeoNet~\cite{Yin18}                                           & $\bullet$ & 10.81              & - \\  
 GLNet~\cite{Chen19}                                          &                & 8.35                & - \\  
 \textbf{RM-Depth (w/o warping)}                         &                & 10.45              & 43.95\% \\      
 \textbf{RM-Depth}                                                &                & \textbf{8.16}   & 31.47\% \\                    
 \thickhline 
\end{tabular}}
\end{table}

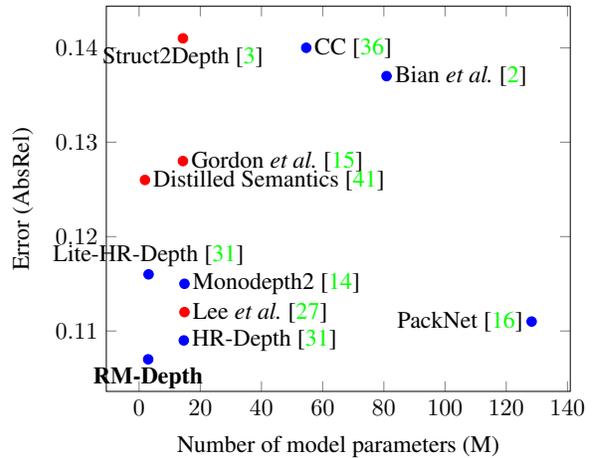
\begin{figure}[t]
\centering
\begin{tikzpicture}[scale=0.9]
\begin{axis}[
    xlabel = Number of model parameters (M),
    ylabel = Error (AbsRel)]
    \addplot[
        scatter/classes={b={blue}, r={red}},
        scatter, mark=*, only marks, 
        scatter src=explicit symbolic,
        nodes near coords*={\Label},
        visualization depends on={value \thisrow{label} \as \Label} ,
        visualization depends on={value \thisrow{anchor}\as\myanchor},
        every node near coord/.append style={anchor=\myanchor},
    ] table [meta=class] {
        x y class label anchor
        14.34 0.141 r {Struct2Depth~\cite{Casser19}} north
        54.64 0.140 b {CC~\cite{Ranjan19}} west
        80.88 0.137 b {Bian \textit{et al.}~\cite{Bian19}} west
        14.33 0.128 r {Gordon \textit{et al.}~\cite{Gordon19}} west
        1.93 0.126 r {Distilled Semantics~\cite{Tosi20}} west
        3.1 0.116 b {Lite-HR-Depth~\cite{Lyu21}} south
        14.84 0.115 b {Monodepth2~\cite{Godard19}} west
        14.84 0.112 r {Lee \textit{et al.}~\cite{Lee21}} west
        14.62 0.109 b {HR-Depth~\cite{Lyu21}} west
        128.29 0.111 b {PackNet~\cite{Guizilini20}} east
         2.97 0.107 b \textbf{RM-Depth} north
    };
\end{axis}
\end{tikzpicture}
\caption{Error of depth models on KITTI against the number of model parameters. Red dots denote models requiring semantics.}\label{fig:depth error vs number of parameters}
\end{figure}

\vspace{+0.1em}\noindent\textbf{Model Size and Runtime.} As shown in Fig.~\ref{fig:depth error vs number of parameters}, RM-Depth just requires 2.97M parameters for the depth model while it outperforms the prior works even for those with semantics. RM-Depth runs at 40FPS for a single depth prediction on a
machine equipped with a GeForce GTX 1080.

\subsection{Ablation Study}

The contributions of the proposed components are studied by evaluating different variants of RM-Depth. Since moving objects are limited on KITTI, the proposed components that are related to object motion are evaluated on Cityscapes. All the results are evaluated on their testing splits and are capped at 80m per standard practice.

\begin{figure}[t]
\captionsetup[subfigure]{labelformat=empty}
\centering
\subfloat[]{\includegraphics[width=2.8cm,trim={8.2cm 0.4cm 4cm 2.0cm},clip]{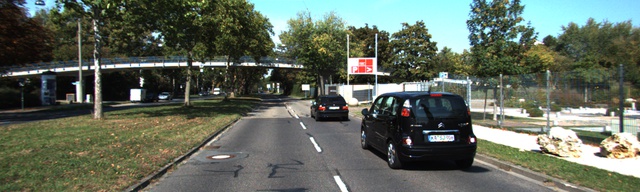}}
\subfloat[]{\includegraphics[width=2.8cm,trim={8.2cm 0.4cm 4cm 2.0cm},clip]{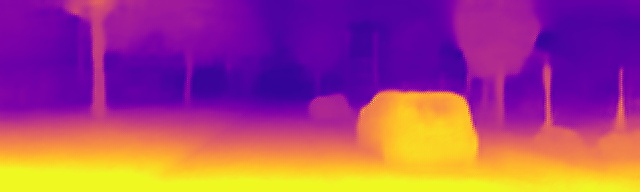}}
\subfloat[]{\includegraphics[width=2.8cm,trim={8.2cm 0.4cm 4cm 2.0cm},clip]{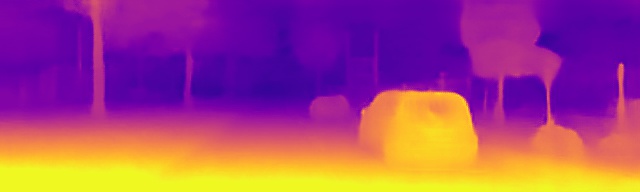}}\\ \vspace{-1.2em}
\subfloat[]{\includegraphics[width=2.8cm,trim={0cm 1.5cm 12.5cm 1cm},clip]{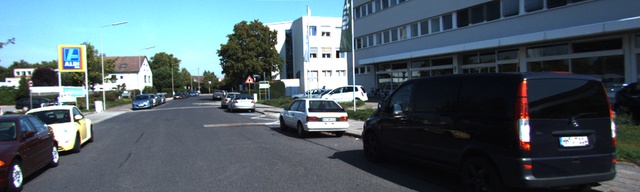}}
\subfloat[]{\includegraphics[width=2.8cm,trim={0cm 1.5cm 12.5cm 1cm},clip]{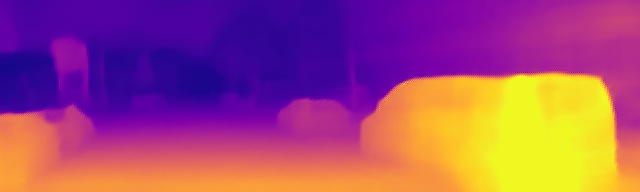}}
\subfloat[]{\includegraphics[width=2.8cm,trim={0cm 1.5cm 12.5cm 1cm},clip]{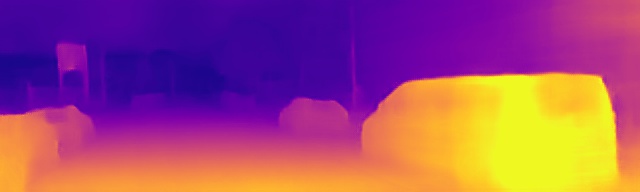}}\\ \vspace{-1.2em}
\subfloat[RGB image]{\includegraphics[width=2.8cm,trim={2cm 2.2cm 10cm 0cm},clip]{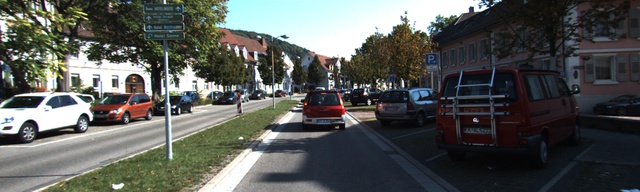}}
\subfloat[Conventional~\cite{Shi16}]{\includegraphics[width=2.8cm,trim={2cm 2.2cm 10cm 0cm},clip]{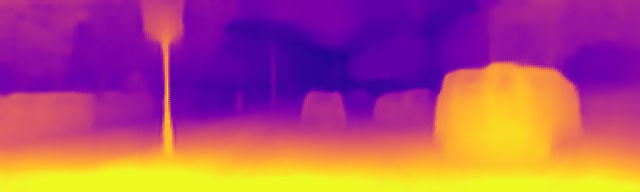}}
\subfloat[Residual upsampling]{\includegraphics[width=2.8cm,trim={2cm 2.2cm 10cm 0cm},clip]{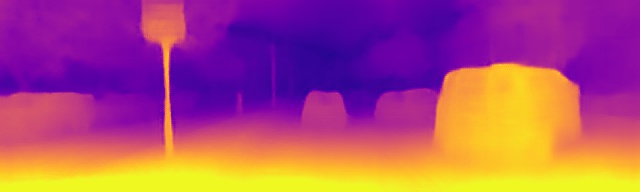}}\\ 
\caption{Depth map predictions using different upsamplings.}\label{fig:depth map using different upsamplings}
\end{figure}

\vspace{+0.1em}\noindent\textbf{RMU and Residual Upsampling.} As shown in Table~\ref{tab:ablation study of RM-Depth on KITTI}, the full model outperforms the baseline by a large margin. The proposed components are effective in improving the depth accuracy. By removing either the residual upsampling or RMU, the depth error is increased. Thanks to the residual upsampling, depth edges are less dispersed comparing to the results using conventional upsampling~\cite{Shi16} as demonstrated in Fig.~\ref{fig:depth map using different upsamplings}. A RMU consists of modulation and update parts. When the modulation part is removed, the depth error is increased. This indicates that the depth improvement is largely benefited by the modulation since it adaptively modifies the feed-forward behavior of the encoder.

\begin{table}[t]
\centering
\caption{Ablation study of RM-Depth on KITTI.}\label{tab:ablation study of RM-Depth on KITTI}
\vspace{-0.5em}
\scalebox{0.72}{
\begin{tabular}{l|cccc} 
 \thickhline
 Model & \multicolumn{4}{c}{Error} \\ 
            & AbsRel & SqRel & RMS & RMSlog \\ 
 \hline
 \textbf{full}                                    &\textbf{0.1081} & \textbf{0.7100} & \textbf{4.5138} & \textbf{0.1831}\\ 
 w/o residual upsampling             & 0.1097 & 0.7313 & 4.5269 & 0.1839\\
 w/o RMU                                      & 0.1167 & 0.8186 & 4.7100 & 0.1895\\
 w/o modulation                            & 0.1165 & 0.7546 & 4.6623 & 0.1910\\
 \hline
 baseline (w/o my contributions) & 0.1187 & 0.8382 & 4.7894 & 0.1927\\
 \thickhline
\end{tabular}}
\end{table}

\vspace{+0.1em}\noindent\textbf{Object Motion.} The full model performs the best among all the variants as summarized in Table~\ref{tab:ablation study of RM-Depth on Cityscapes}. The proposed components are effective in improving depth accuracy on non-rigid scenes. When warping is disabled, the source images are not warped towards the target image. There is a large ``visual gap'' between images in the pair, and hence the depth error is increased. By disabling the outlier-aware regularization, the depth accuracy is deteriorated. Comparing to the variant using the sparsity loss~\cite{Li20}, the full model performs much better. When object motion estimation and other proposed components are disabled, it has been experienced that the training becomes diverged after 6 epochs. Holes often appear on moving objects as revealed in Fig.~\ref{fig:depth predictions on Cityscapes}.

\begin{table}[t]
\centering
\caption{Ablation study of RM-Depth on Cityscapes.}\label{tab:ablation study of RM-Depth on Cityscapes}
\vspace{-0.5em}
\scalebox{0.72}{
\begin{tabular}{l|cccc} 
 \thickhline
 Model & \multicolumn{4}{c}{Error} \\ 
            & AbsRel & SqRel & RMS & RMSlog \\ 
 \hline
 \textbf{full}                                        & \textbf{0.0903} & \textbf{0.8248} & \textbf{5.5027} & \textbf{0.1430}\\ 
 w/o warping                                      & 0.0933 & 0.9248 & 5.6283 & 0.1461\\
 w/o outlier-aware regularization      & 0.0995 & 0.9986 & 5.8281 & 0.1545\\
 using sparsity loss as~\cite{Li20}  & 0.1066 & 1.1073 & 6.0965 & 0.1642\\
 w/o object motion estimation         & 0.1174 & 1.1195 & 6.4542 & 0.1729\\
 \hline 
 baseline (w/o my contributions)     & 0.1335 & 1.8784 & 6.9748 & 0.1912\\
 \thickhline
\end{tabular}}
\end{table}

\vspace{+0.1em}\noindent\textbf{Number of RMUs.} Compromising accuracy and computational complexity, at most 2 RMUs are assigned for levels 2 – 3. As summarized in Table~\ref{tab:ablation study of the number of RMUs on KITTI}, depth accuracy and runtime increase with the number of RMUs.

\begin{table}[t]
\centering
\caption{Ablation study of the number of RMUs on KITTI.}\label{tab:ablation study of the number of RMUs on KITTI}
\vspace{-0.5em}
\scalebox{0.72}{
\begin{tabular}{l|cccc|c} 
 \thickhline
 Number of RMUs & \multicolumn{4}{c|}{Error} & Runtime\\ 
                               & AbsRel & SqRel & RMS & RMSlog          & [ms]\\ 
 \hline
 3 (L4: 1, L3: 1, L2: 1)                & 0.1161 & 0.7713 & 4.6799 & 0.1906 & \textbf{14.99}\\ 
 6 (L4: 2, L3: 2, L2: 2)                & 0.1135 & 0.7490 & 4.6128 & 0.1877 & 20.40\\ 
 8 (L4: 4, L3: 2, L2: 2)                & 0.1098 & 0.7251 & 4.5535 & 0.1845 & 22.07\\ 
 \textbf{13 (L4: 9, L3: 2, L2: 2)} & \textbf{0.1081} & \textbf{0.7100} & \textbf{4.5138} & \textbf{0.1831} & 24.78\\ 
 \thickhline
\end{tabular}}
\end{table}

\section{Conclusion}

RM-Depth, an unsupervised learning framework, is proposed for single-image depth estimation. Complete motion that includes camera and object motions is used to assist the unsupervised learning. This breaks down the scene rigidity requirement. The depth network utilizes recurrent modulation units for dynamic and iterative feature fusion. The use of residual upsampling enables specific upsampling of different spectral components. For the motion network, a warping-based approach has been devised to recover object motion. An outlier-aware regularization loss has also been exploited. With the proposed innovations, the depth network achieves promising results while it only requires 2.97M model parameters.

{\small
\bibliographystyle{ieee_fullname}
\bibliography{egbib}
}

\end{document}